\documentclass[letterpaper]{article}
\usepackage{ijcai09}

\usepackage{times}
\usepackage{amssymb}
\usepackage{amsthm}
\usepackage{amsmath}
\usepackage{amsfonts}
\usepackage{graphicx}
\usepackage{graphicx}\graphicspath{{figures/}}
\usepackage{subfigure}
\usepackage{wrapfig}
\usepackage{floatflt}
\usepackage{appendix}
\usepackage{color}
\usepackage{epsfig}
\usepackage{array}

\definecolor{MyGray}{rgb}{0.9,0.9,0.9}


\newcommand{\norm}[1]{\left\vert\left\vert #1 \right\vert\right\vert}
\newcommand{\Ep}{\mathbb{E}}

\title{Exponential Family Hybrid Semi-Supervised Learning}

\author{Arvind Agarwal \\
School of Computing\\
University of Utah \\
arvind@cs.utah.edu
\And
Hal Daum\'e III\\
School of Computing \\
University of Utah\\
hal@cs.utah.edu\\
}

\newcommand{\secref}[1]{Section~\ref{sec:#1}}
\newcommand{\figref}[1]{Figure~\ref{fig:#1}}
\newcommand{\tabref}[1]{Table~\ref{tab:#1}}
\newcommand{\pd}[2]{\frac{\partial #1}{\partial #2}}
\newcommand{\comment}[1]{}

\def\th{{\theta}}
\def\tht{{\tilde{\theta}}}
\def\x{{\mathbf{x}}}

\begin{document}

\maketitle

\begin{abstract}
We present an approach to semi-supervised learning based on an exponential family characterization.  Our approach generalizes previous work on coupled priors for hybrid generative/discriminative models.  Our model is more flexible and natural than previous approaches.  Experimental results on several data sets show that our approach also performs better in practice.
\end{abstract}

\section{Introduction}
\label{sec:introduction}

Labeled data on which to train machine learning algorithms is often scarce or expensive.  This has led to significant interest in semi-supervised learning methods that can take advantage of unlabeled data \cite{Cozman03,zhu05survey}.  While it is straightforward to integrate unlabeled data in a generative learning framework \cite{Nigam2000}, it is not so in a discriminative framework.  Unfortunately, it is well-known both empirically and theoretically \cite{Ng2002} that discriminative approaches tend to outperform generative approaches when there is enough labeled data.  This has led to many recent developments in \emph{hybrid} generative/discriminative models that are able to leverage the power of both frameworks (see \secref{relatedwork}).  One particular such example is the work of Lasserre et al.~\shortcite{Minka2006}, who describe a hybrid framework (``PCP'') in which a generative model and discriminative model are jointly estimated, using a \emph{prior} that encourages them to have similar parameters.

In this paper, we generalize the {\it parameter coupling prior}\footnote{Terms {\it parameter coupling prior} and {\it coupled prior} were introduced in \cite{Druck2007} and do not appear in \cite{Minka2006} though they refer to the framework introduced in \cite{Minka2006}.} (PCP) method \cite{Minka2006} to arbitrary distributions belonging to the {\it exponential family}.  Unlike the PCP method, we do not restrict ourselves to the {\it Gaussian prior}, but instead choose a prior that is natural to the model.  Other authors \cite{Bouchard2007} have also noted the inappropriateness of the Gaussian prior to couple the generative and discriminative models.  Our resulting approach for hybridizing discriminative and generative models is: (1) not restricted to a particular class of the models; (2) more flexible in choosing the way these two models can be combined; (3) enables us to achieve a closed form solution for the generative parameters, unlike PCP method, where one has to resort to the numerical optimization.  We demonstrate our framework on using a Beta/Binomial conjugate pair on the text categorization problems addressed by Druck et al.~\shortcite{Druck2007}.

\section{Background}
\label{sec:background}

In general, machine learning approaches to classification can be divided into two categories: {\it generative approaches} and {\it discriminative approaches}.  Generative approaches assume that the data is generated though an underlying process.  One simple example is document categorization: for each example $(\x,y)$, we first choose a category $y$, and then produce a document $\x$ conditioned on the category $y$.  The goal in generative modeling is to approximate the joint distribution $p(\x,y)$ that represents this process.  On the other hand, discriminative approaches do not assume any underlying process and directly model the probability of category given the document $p(y|\x)$. Ng and Jordan~\shortcite{Ng2002} compare these two approaches and show that while discriminative models are asymptotically better than generative models, generative models need less data to train.

In semi-supervised settings, one has access to lots of unlabeled data but only a small amount of labeled data. It is easy to see that unlabeled data is not directly useful in a discriminative setting but can be easily used in generative setting.  However, since discriminative methods asymptotically tend to outperform generative methods \cite{Ng2002}, this naturally leads to combining these two approaches and building a hybrid model that does better than the individual models.  Earlier work \cite{Triggs2004,Minka2006,Druck2007} has shown the efficacy of the hybrid approach.

\subsection{Exponential Family and Conjugate Priors}
\label{sec:exponential}
For the sake of completeness, we briefly define the exponential family which we will use as the basis of our hybrid model. The exponential family is a set of distributions whose probability density function\footnote{``Density'' can be replaced by ``mass'' in the case of discrete random variable.} can be expressed in the following form:
\begin{equation}
f(\x;\th)=h(\x)\exp(\langle \eta(\theta)T(\x) \rangle -A(\th))
\label{expdist}
\end{equation}
Here $T(\x)$ is sufficient statistics, $\eta{(\theta)}$ is a function of natural parameters $\theta$, and $A(\theta)$ is a normalization constant (also known as {\it log-partition function}).  


One important property of the exponential family is the existence of conjugate priors.  Given any member of the exponential family in Eq~\eqref{expdist}, the \emph{conjugate prior} is a distribution over its \emph{parameters} with the following form:
\begin{equation}
p(\th|\alpha,\beta)=m(\alpha,\beta) \;\exp(\langle \eta(\th), \alpha \rangle - \beta A(\th))
\label{conjprior}
\nonumber
\end{equation}
Here $\langle a,b \rangle$ denotes the dot product of vectors $a$ and $b$.  Both $\alpha$ and $\beta$ are hyperparameters of the conjugate prior.  Importantly, function $A(\cdot)$ is the same between the exponential family member and the conjugate prior.

A second important property of the exponential family is the relationship between the log-partition function $A(\th)$ and the sufficient statistics.  In particular, we have:
\begin{equation}
\label{expfamnorm}
\frac {\partial A}{\partial \th} = \Ep_\th \left[ T(\x) \right]
\end{equation}

\subsection{Hybrid Model with Coupled Prior}
\label{sec:hybrid}
We first define the problem and some of the notations that we will use through-out the paper. Our task is to learn a model that predicts a label $y$ given an example $\x$.  We are given the data $D=D_L \cup D_U$ where $D_L$ represents the labeled data and $D_U$ represents the unlabeled data.  Each instance of the labeled data consists of a pair $(\x,y)$ where $\x$ is feature vector and $y$ is the corresponding label.  Each instance of unlabeled data consists of only feature vector $\x$. The $\x$s are $M$-dimensional feature vectors, and $\x_d$ denotes the $d^{\text{th}}$ feature.

We now give a brief overview of the hybrid model presented by Lasserre et al.~\shortcite{Minka2006}. The hybrid model is a mixture of discriminative and generative components, both of which have separate sets of parameters. These two sets of parameters (hence two models) are combined using a prior called {\it coupled prior}.  Considering only one data point (the extension to multiple data points is straightforward and presented later), the model is defined as follows:
\begin{eqnarray}
\nonumber
p(\x,y,\th,\tht)&=&p(\tht,\th)p(y|\x,\th)p(\x|\tht) \\
\nonumber
&=&p(\tht,\th)p(y|\x,\th)\sum_{y'}p(\x,y'|\tht)
\end{eqnarray}
Here $\th$ is a set of discriminative parameters, $\tht$ a set of generative parameters, and $p(\tht,\th)$ provides the natural coupling between these two sets of parameters.  $p(y|\x,\th)$ is the discriminative component; $p(\x|\tht) = \sum_{y'}p(\x,y'|\tht)$ is the generative component.

The most important aspect of this model is the \emph{coupled prior} $p(\tht,\th)$, which interpolates the hybrid model between two extremes; generative model when $\th=\tht$ and discriminative when $\th$ is independent of $\tht$. In other cases, the goal of the coupled prior is to encourage the generative model and the discriminative model to have similar parameters.  In earlier work \cite{Minka2006,Druck2007}, a Gaussian prior was used as the coupled prior:
\begin{equation}
p(\tht,\th) \varpropto \exp\left[ -\frac 1 {2\sigma^2} \norm{\tht - \th}^2 \right]
\nonumber
\end{equation}
Unfortunately, the Gaussian prior is not always appropriate \cite{Bouchard2007}.

\section{Exponential Family Hybrid Model}

In this section, we provide a more general prior for the hybrid model that is not only mathematically convenient but also allows choosing a problem specific prior.

\subsection{Exponential Family Generalization}

First, we generalize the hybrid model defined in \secref{hybrid} for the distributions that come from the exponential family.  In other words, all of the distributions (generative, discriminative and coupled prior) of the generalized hybrid model belong to the exponential family.  We first provide the definitions of discriminative and generative models in terms of exponential family.

\vspace{1em}
\noindent
{\bf Generative model:}
\begin{equation}
p(\x,y|\tht)=h(\x,y)\exp(\langle\tht, T(\x,y)\rangle-A(\tht))
\label{expgen}
\end{equation}
{\bf Discriminative model:}
\begin{equation}
p(y|\x,\th)=g(y)\exp(\langle \th, T(\x,y) \rangle -B(\th,\x))
\label{expdisc}
\end{equation}

Next, we break the coupled prior $p(\tht,\th)$ into two parts; an independent prior on the discriminative parameters $p(\th)$ and a prior on the generative parameters given discriminative parameters $p(\tht|\th)$. This formulation lets us model the dependency of the generative component over the discriminative component.  Our new hybrid model is now defined as:
{\small
\begin{equation}
p(\x,y,\th,\tht)=\Big[ p(\th)p(y|\x,\th)\Big]\; p(\tht|\th) \; \Big[ \sum_{y'}p(\x,y'|\tht)\Big]
\label{rawmodel}
\end{equation}
}
For convenience and interpretability (later we will show that it also improves the performance), we choose the coupled prior $p(\tht|\th)$ to be conjugate with the generative model.

\vspace{1em}
\noindent
{\bf Conjugate prior:}
\begin{equation}
p(\tht|\th)=m(\th) \;\exp(\langle \tht, \alpha(\th) \rangle - \beta(\th) A(\tht))
\label{expprior}
\end{equation}
Here, $\alpha(\cdot)$ and $\beta(\cdot)$ are user-defined functions that map the discriminative parameters $\th$ into hyperparameters for the conjugate prior.  We discuss suitable choices of these functions in Section~\ref{sec:practicePrior}.

Substituting the exponential definitions of generative model Eq~\eqref{expgen}, discriminative model Eq~\eqref{expdisc}, and coupled prior Eq~\eqref{expprior} in Eq~\eqref{rawmodel}, and taking a log, we obtain a log joint probability of data and parameters:
\begin{eqnarray}
\label{loglik}
&&\hspace{-2em}L = \log p(\x,y,\th,\tht)= \\
&& \log p(\th) + \nonumber\\
&& \log m(\th) + \langle \tht \alpha(\th) \rangle - \beta(\th) A(\tht) + \nonumber\\
&& \log g(y) + \sum_{(\x,y) \in D_L} \Big[ \langle \theta, T(\x,y) \rangle - B(\th, \x) \Big] +  \nonumber\\
&& \quad\quad \sum_{\x\in D} \log  \sum_{y'} \Big[ h(\x,y') \exp( \langle\tht, T(\x,y') \rangle - A(\tht)) \Big]\nonumber
\end{eqnarray}
Note that here discriminative part is defined only for labeled data while generative part is defined for both labeled and unlabeled data.

\subsection{Parameter Optimization}
We perform parameter optimization by a coordinate descent method, alternating between optimizing the discriminative parameters $\th$ and optimizing the generative parameters $\tht$.

For the generative parameters, we take the partial derivative of the log probability in Eq~\eqref{loglik} with respect to $\tht$:
{\small
\begin{eqnarray}
\nonumber \pd{L}{\tht}&=& \alpha(\th) - \beta(\th) A'(\tht) +  \\
\nonumber &&\sum_{\x\in D}\sum_{y'}p(y'|\x,\tht)(T(\x, y')-A'(\tht))
\end{eqnarray}
}
Here, $p(y'|\x,\tht)$ is the probability based on the parameters estimated in the last iteration $p(y'|\x,\tht_{old})$. Substituting this in the above equation and setting it equal to zero, we obtain:
{\small
\begin{eqnarray}
\nonumber A'(\tht)&=&\frac{\sum_{\x\in D}\sum_{y'}p(y'|\x,\tht_{old})T(\x, y')+\alpha(\th)}{N+\beta(\th)}\\
	&=&\frac{\hat{\Ep}_{\x\sim D}\,\Ep_{y\sim \tht_{old}}(T(\x, y'))+\alpha(\th)}{N+\beta(\th)}
\label{delexpgen}
\end{eqnarray}
}
Here $A'(\tht)$ denotes the partial derivative of $A(\tht)$ with respect to $\tht$. As discussed in \secref{introduction}, choosing a conjugate prior gives us a closed form solution for $A'(\tht)$.  From Eq~\eqref{expfamnorm}, we know that $A'(\tht)$ is equivalent to the expected sufficient statistics of the generative model.

Having solved for the generative parameters $\tht$, we now solve the hybrid model for discriminative parameters $\th$.
{\small
\begin{eqnarray}
\nonumber \pd{L}{\th}&=& \pd{\log p(\th)}{\th} + \pd{\log m(\th)}{\th} + \tht \alpha'(\th) -   \\
&& \beta'(\th) A(\tht) + \sum_{(\x,y)\in D_L} (T(y,\x)-B'(\th,\x))
\label{delexpdisc}
\end{eqnarray}
}
There is no closed form solution to the above expression therefore we solve it using numerical methods. In our implementation, we use stochastic gradient descent.

\subsection{Hybrid Multiple Binomial Model}
\label{sec:practice}

In this section, we see how this hybrid model can be applied in practice.  We first choose a generative model that is suitable to our application.  We next choose the coupled prior conjugate to the generative model.  Since later on, we intend to use the hybrid model for the document classification task, we use a {\it naive Bayes}\footnote{It should be noted that ``naive Bayes'' classifiers come into (at least) two different versions: the ``multivariate Bernoulli version'' and the ``multinomial version'' \cite{Mccallum1998}. Because of its generality, in our implementation, we use multivariate Bernoulli.} (NB) model for the generative part and {\it logistic regression} for the discriminative part, akin to the study of Ng and Jordan~\shortcite{Ng2002}. The generative part of our model (naive Bayes) is given by:
\begin{eqnarray}
\nonumber p(y,\x|\pi,v)&=&p(y|\pi)p(\x|y,v) \\
&=& \prod_k \pi_k^{1_{\{y=k\}}} \prod_d v_{yd}^{\x_d}(1-v_{yd})^{1-\x_d}
\label{nb}
\end{eqnarray}
Here, $\sum_{y'}\pi_{y'}=1$ and $0\le v_{yd}\le 1$. $1_{\{y=k\}}$ is an indicator function that takes value $1$ if $y=k$ and $0$ otherwise. 

The discriminative part is:
\begin{eqnarray}
p(y|\x,w,b) &=& \frac{1}{Z_{\x}}\exp\Big(b_y + \sum_d\x_dw_{yd}\Big)
\label{lr}
\end{eqnarray}
Where $Z_{\x}=\sum_{y'}\exp\big(b_{y'} + \sum_d \x_d w_{y'd}\big)$ is a normalization constant. Note here that since these models form generative/discriminative pair, number of parameters is same in both models.  It is easy to see that there is one-to-one relationship between these two sets of parameters.  $b_y$ in the discriminative model behaves similar to $\pi_y$ in the generative model, and $w_{yd}$ behaves similar to $v_{yd}$.  Since $w_{yd}$ and $v_{yd}$ are the parameters that capture most of the information, we use coupled prior to couple these sets of parameters and do not couple $b_y$ and $\pi_y$. It is important to note the difference between the canonical parameters of the exponential family representation of the model and the mean parameters. In the generative(or discriminative) model, $\tht_{yd}$ (or $\th_{yd}$) denote the canonical parameters while $v_{yd}$ (or $w_{yd}$) denote the mean parameters. 

Having defined the appropriate discriminative and generative models, now we can get equivalent exponential family forms of these models.  First we show the exponential form of the generative model. The generative model in Eq~\eqref{nb} can be broken into two parts: one is class probability $p(y|\pi)$ and other class conditional probability $p(\x|y,v)$.  Since the parameters of these distributions are independent, we can get their exponential representations separately.  Considering the class conditional probability for one feature, Eq~\eqref{nb} can be written in the following form:
{\small
\begin{align}
\nonumber p(\x_d|y,v_{yd}) = \exp\left(\x_d \log \frac{v_{yd}}{1-v_{yd}} + \log (1-v_{yd})\right)
\end{align}
}
Comparing this with Eq~\eqref{expgen} gives $\tht_{yd}=\log \frac{v_{yd}}{1-v_{yd}}$; $A(\tht_{yd}) = \log(1+e^{\tht_{yd}})$ and $T(y,\x)=\x_d$. Substituting these along with the appropriate conjugate prior in Eq~\eqref{delexpgen} gives us a closed form solution for $A'(\tht_{yd})$, which, in the naive Bayes model is equal to $v_{yd}$.
{\small
\begin{eqnarray}
A'(\tht_{yd})=v_{yd}= \frac{\sum_{\x \in D}p(y|\x,\tht_{old})\x_d + \alpha(\th)}{N+\beta(\th)}
\label{delnb}
\end{eqnarray}
}
In other words, $v_{yd}$ is the normalized expected count of the $d_{th}$ feature in class $y$, with smoothing parameters that are controlled by the coupled prior hyperparameters $\alpha(\th)$ and $\beta(\th)$.

Next we solve for $\pi_y$ by directly optimizing the objective function Eq~\eqref{nb} with respect to $\pi_y$ with the given constraints. This gives us $\pi_y=\frac{\sum_{\x \in D}p(y|\x,\tht_{old})}{N}$  which is the normalized expected number of examples in class $y$.

Having solved for generative parameters, we now solve for the discriminative parameters. Ideally, we would like to first get an equivalent exponential form of Eq~\eqref{lr} and then solve it using Eq~\eqref{delexpdisc}. Since Eq~\eqref{delexpdisc} is only defined for discriminative parameters that are coupled ($w$), we can not use Eq~\eqref{delexpdisc} unless we break Eq~\eqref{lr} into two exponential forms separate for $w$ and $b$ and, it is not clear how to do so. Therefore, we solve for discriminative parameters directly, without converting Eq~\eqref{lr} into exponential form. It is important to note here that mean parameters $w$ in Eq~\eqref{lr} is equal to the canonical parameters $\th_{yd}$. We place Gaussian prior $p(\th) = N(\th | 0,\sigma^2)$ on $w=\th$ and an improper uniform prior on $b$.  Taking derivatives, we obtain:
{\footnotesize
\begin{align*}
\pd{L}{w_{yd}} &= -\frac{w}{\sigma^2} + \pd{\log m(w_{yd})}{w_{yd}} + \langle \tht_{yd} \alpha'(w_{yd}) \rangle - \beta'(w_{yd})A(\tht_{yd})  \nonumber\\
  & + \sum_{(\x,y')\in D_L}\Big\{ 1_{\{\x_d=1\}}   - \frac{1}{Z_{\x}}\exp(b_{y'} + \sum_d \x_d w_{y'd}) 1_{\{\x_d=1\}} \Big\} \\
\pd{L}{b_y} & =  \sum_{(\x,y')\in D_L}\Big\{ 1_{\{y=y'\}} - \frac{1}{Z_{\x}}\exp(b_y + \sum_d\x_dw_{yd}) \Big\} 
\end{align*}
}
\subsection{Conjugate Beta Prior}
\label{sec:practicePrior}


Recall that our conjugate prior crucially depends on two functions: $\alpha(\th)$ and $\beta(\th)$ that ``convert'' the discriminative parameters $\th$ into a prior on the generative parameters $p(\tht | \th)$.  In the case of the binomial likelihood, the conjugate prior is Beta. Exponential form of Beta prior is defined as:
\begin{align}
p(\tht_{yd}|\th_{yd})= m(\th_{yd}) \exp(\tht_{yd} \alpha(\th_{yd}) - \beta(\th_{yd}) A(\tht_{yd})) 
\label{betaprior}
\nonumber
\end{align}
Where $m(\th_{yd})=\frac{\Gamma(\beta(\tht_{yd})+2)}{\Gamma(\alpha(\tht_{yd})+1)\Gamma(\beta(\th_{yd})-\alpha(\th_{yd})+1)}$ and $A(\tht_{yd} = \log(1+e^{\tht_{yd}})$.

\begin{figure}[tc]
\begin{center}
\includegraphics[width=0.50\textwidth]{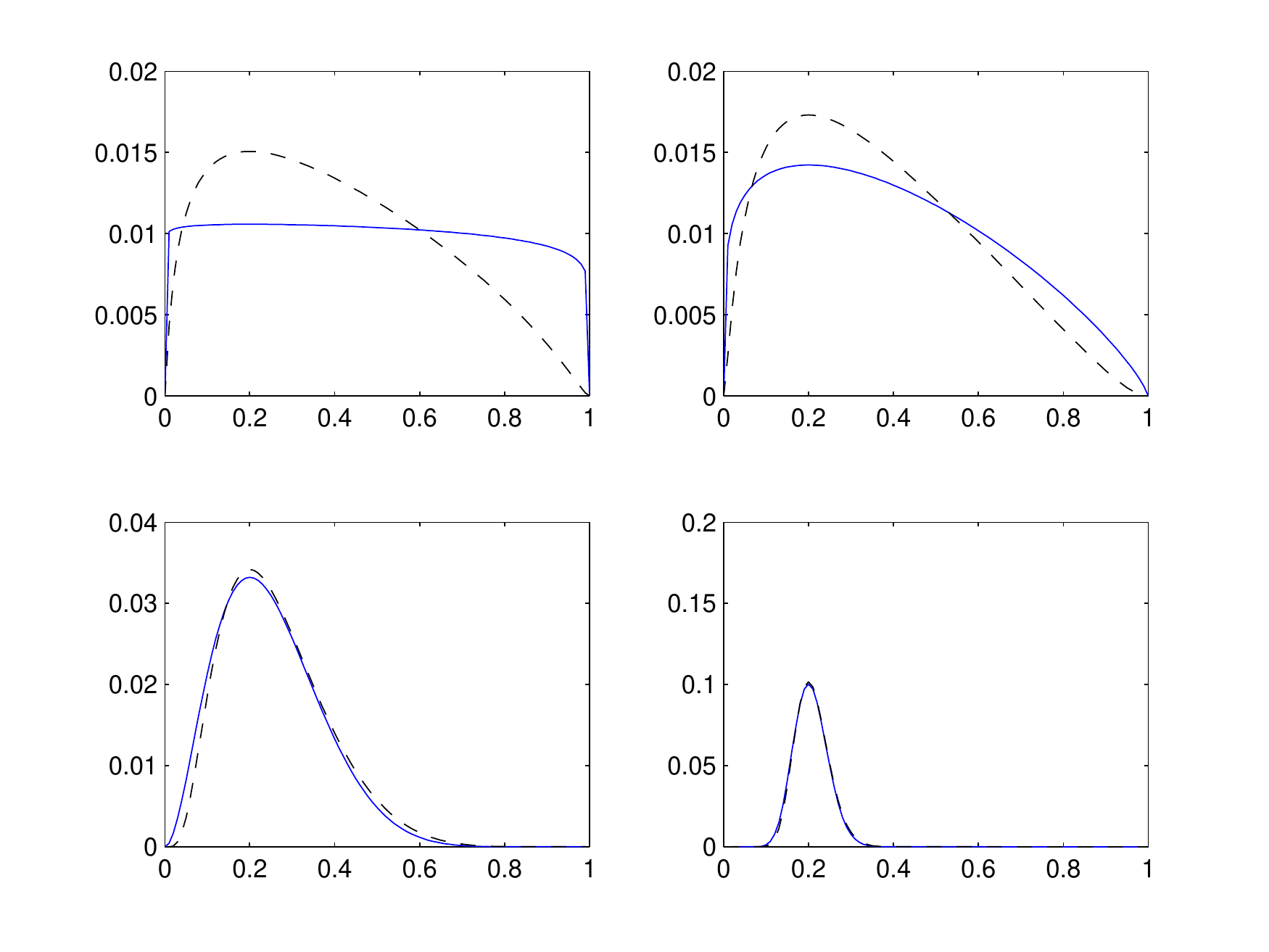}
\caption{Effect of gamma on the Beta prior (solid curve) and logistic-Normal prior (dashed curve) for gamma=0.1, 1, 10, 100 (top-left, top-right, bottom-left, bottom-right) and for the transformed discriminative parameter $T(w)=0.2$}
\label{fig:betadist}
\end{center}
\end{figure}

We select the function $\alpha(\th_{yd})$ and $\beta(\th_{yd})$ to be such that: (1) the \emph{mode} of the conjugate prior is $\th_{yd}$ and (2) the \emph{variance} of the conjugate prior is controllable by the hyperparameter $\gamma$.  As noted from Figure~\ref{fig:betadist}, as $\gamma$ goes to $\infty$, variance goes to $0$ and prior forces generative parameters to be equal to the discriminative parameters (pure generative model) and as $\gamma$ goes to $0$, variance goes to $\infty$ which implies the independence between generative and discriminative parameters (pure discriminative model). Other values of $\gamma$ interpolate between these two extremes. Thus, we choose $\alpha(\th_{yd}) = \gamma / (1 + e^{-\th_{yd}})$ and $\beta(\th_{yd}) = \gamma$.  This gives mode of $p(\tht_{yd}|\th_{yd})$ at $\th_{yd}$ with the variance that decreases in $\gamma$, as desired.

It is important to note that our choice of hyperparameters for the conjugate prior is not specific to this example, but holds true in general.  In the general case, let $A$ be the log-partition function associated with the generative model, then, the conjugate prior hyperparameters should be $\alpha(\th) = \gamma A'(\th)$ and $\beta(\th)=\gamma$. This gives us the mode of conjugate prior at $\th$ with the variance that decreases in $\gamma$. In the beta/binomial hybrid model, $A'(\th)=A'(w)=1/(1+e^{-w})$. Also note that in the beta/binomial example, $A'(\th)$ is also the transformation function $T$ that transforms the discriminative mean parameters $w$ to the generative mean parameters $v$.


In Figure~\ref{fig:betadist}, we also compare the Beta prior (solid blue curves) to an ``equivalent'' logistic-Normal prior (dashed black curves) for four settings of $\gamma$.  The logistic-Normal is parameterized to have the same mode and variance as the Beta prior.  As we can see, for high values of $\gamma$ (wherein the model is essentially generative), the two behave quite similarly.  However, for more moderate settings of $\gamma$, the priors are qualitatively quite different.

%

\section{Related Work}
\label{sec:relatedwork}
There have been a number of efforts to combine generative and discriminative models to obtain a hybrid model that performs better than either individually.  Some of the earlier works \cite{Raina2003,Triggs2004} use completely different approaches to hybridize these models; Raina et al.~\shortcite{Raina2003} present a model for the document classification task where a document is split into multiple regions and complementary properties of generative/discriminative models are exploited by training a large set of the parameters generatively and only a small set of parameters discriminatively.  Bouchard and Triggs~\shortcite{Triggs2004} build a hybrid model by taking a linear combination of generative and discriminative model.  This model is similar to the multi-conditional learning model presented by McCallum et al.~\shortcite{Mccallum06}.  Jaakkola and Haussler~\shortcite{Jaakkola99} describe a scheme in which the kernel of a discriminative classifier is extracted from a generative model.  Though these models have shown to perform better than just the discriminative or generative model, none of them combine the hybrid model in natural way.  

Our work builds on the work of Lasserre et al.~\shortcite{Minka2006} and Druck et al.~\shortcite{Druck2007}, which are discussed in Section~\ref{sec:hybrid}.  Along these lines, Fujino et al.~\shortcite{Fujino2008} present another hybrid approach where a generative model is trained using a small number of labeled examples.  Since the generative model has high bias, a generative ``bias-correction'' model is trained in a discriminative manner to discriminatively combine the bias-correction model with the generative model.  Most of these work focus on the application and little on the theory of the hybrid model.  There has been a recent work by Bouchard~\shortcite{Bouchard2007} that presents a unified framework for the ``PCP'' model and the ``convex-combination'' model \cite{Triggs2004}, and proves performance properties. 


\begin{table}
\begin{center}
\small
\begin{tabular}{|l| c@{\hspace*{1.7ex}} | m{0.58\linewidth} |}
\hline
 	& {\bf No. of} &  \\
{\bf Dataset} & {\bf Features} &  \multicolumn{1}{|c|}{{\bf Dataset description}} \\
\hline
movie  & $24,841$ & classifies the sentiments of the review of the movies from IMDB as {\it positive} or {\it negative} \\
\hline
webkb  & $22,824$ & classifies webpages from university as {\it student, course, faculty} or {\it project} \\
\hline
sraa  & $77,494$ & classifies messages by the newsgroup to which they were posted: {\it simulated-aviation}, {\it real-aviation}, {\it simulated-autoracing}, {\it real-autoracing} \\
\hline
\end{tabular}
\end{center}
\caption{Description of the datasets used in the experiments}
\label{tab:datasets}
\end{table}

\section{Experiments}
\subsection{Experimental Setup}
In this section, we show empirical results of our approach and compare them with the existing (and most related to our method) state-of-the-art semi-supervised methods \cite{Druck2007}.  In order to have a fair comparison, we use experimental setup of Druck et al.~\shortcite{Druck2007} and perform experiments only for the datasets where PCP model have shown to perform best, There are three such datasets: {\it movie, sraa} and {\it webkb}. Description of these datasets is given in \tabref{datasets}. 

Although all of the examples in these datasets are labeled, we perform experiments by taking a subset of dataset as labeled and treating the rest of the examples as unlabeled.  We use either $10$ or $25$ labeled examples from each class and vary unlabeled examples from $0$ to a maximum of $1000$.  Number of unlabeled examples are same in each class. We show our results for two sets of experiments: (1) we show how performance varies as we vary the number of unlabeled examples; (2) we show how performance varies with respect to $\lambda$. Here $\lambda$ normalizes the $\gamma \in [\infty,0]$ in the range of $[0,1]$ using $\gamma=((1-\lambda)/\lambda)^2$. Now $\lambda=0$ corresponds to the pure generative case while $\lambda=1$ corresponds to the pure discriminative case. As in the work of Druck et al.~\shortcite{Druck2007}, the success of the semi-supervised learning depends on the quality of the labeled examples, therefore we choose five random labeled sets and report the average on them.  In our results, we report the percentage classification accuracy which is the ratio of number of examples correctly classified to the total number of test examples.

\subsection{Results and Discussion}
Results on the above mentioned three datasets are presented in \tabref{compare}. Table shows the results for the PCP model with the Gaussian prior (PCP-Gauss) and with the Beta prior (PCP-Beta). Since PCP-Beta uses the binomial version of NB, we reimplemented the PCP-Gaussfor the binomial version of NB and compare the results with it. Though we also show the results for PCP-Gauss multinomial \cite{Druck2007}, a fair comparison would be to compare only binomial models. \%change is the change in PCP-Beta with respect to the PCP-Gauss binomial version.  As we see, PCP-Beta performs better than PCP-Gauss binomial in all experiments and better than PCP-Gauss multinomial in all experiments except sraa(10). Compared to PCP-Gauss binomial, PCP-Beta performs significantly better on sraa and movie datasets.

\begin{table}
\begin{center}
\footnotesize
\begin{tabular}{|l| c | c | c | c |}
\hline
 & pcp-Gauss &  pcp-Gauss & pcp-Beta &  \%\\
\multicolumn{1}{|c|}{Dataset}  & Mult &  Bin & Bin &  change \\
\hline
movie (10) & $64.6$ & $63.4$ {\tiny$(3.2)$} & $\mathbf{68.3}$ {\tiny$(5.5)$} & $+7.7\%$ \\
movie (25) & $68.6$ & $69.0$ {\tiny$(1.5)$} & $\mathbf{76.7}$ {\tiny$(1.2)$} & $+11.1\%$ \\
webkb (10) & $72.5$ & $73.7$ {\tiny$(3.7)$} & $\mathbf{75.3}$ {\tiny$(2.9)$} & $+2.2\%$ \\
webkb (25) & $76.7$ & $83.8$ {\tiny$(1.3)$} & $\mathbf{83.9}$ {\tiny$(1.6)$} & $+1.1\%$ \\
sraa (10) & $\mathbf{81.6}$ & $67.7$ {\tiny$(6.8)$} & $79.1$ {\tiny$(4.0)$} & $+16.8\%$  \\
sraa (25) & $84.1$ & $76.6$ {\tiny$(3.5)$} & $\mathbf{86.1}$ {\tiny$(1.0)$} & $+12.4\%$ \\
\hline
\end{tabular}
\end{center}
\caption{Comparative results for pcp with Gaussian prior and pcp with Beta prior. Parenthesized values denote the number of labeled examples per class and the standard deviation.}
\label{tab:compare}
\end{table}

Comparing multinomial and binomial versions of PCP-Gauss, we see that for movie and webkbb datasets, binomial version performs better (or almost equal) than the multinomial while for sraa dataset, multinomial performs better. We conjecture that reason for this behavior could be because sraa has a large number of features and feature independence assumption is less violated in multinomial NB than in binomial NB. When datasets do not have too many features, binomial version tends to perform better because binomial NB accounts for both presence and absence of the features, in contrast to multinomial NB which only accounts for the presence of the features.

\begin{figure}[tc]
\begin{center}
\includegraphics[width=0.5\textwidth]{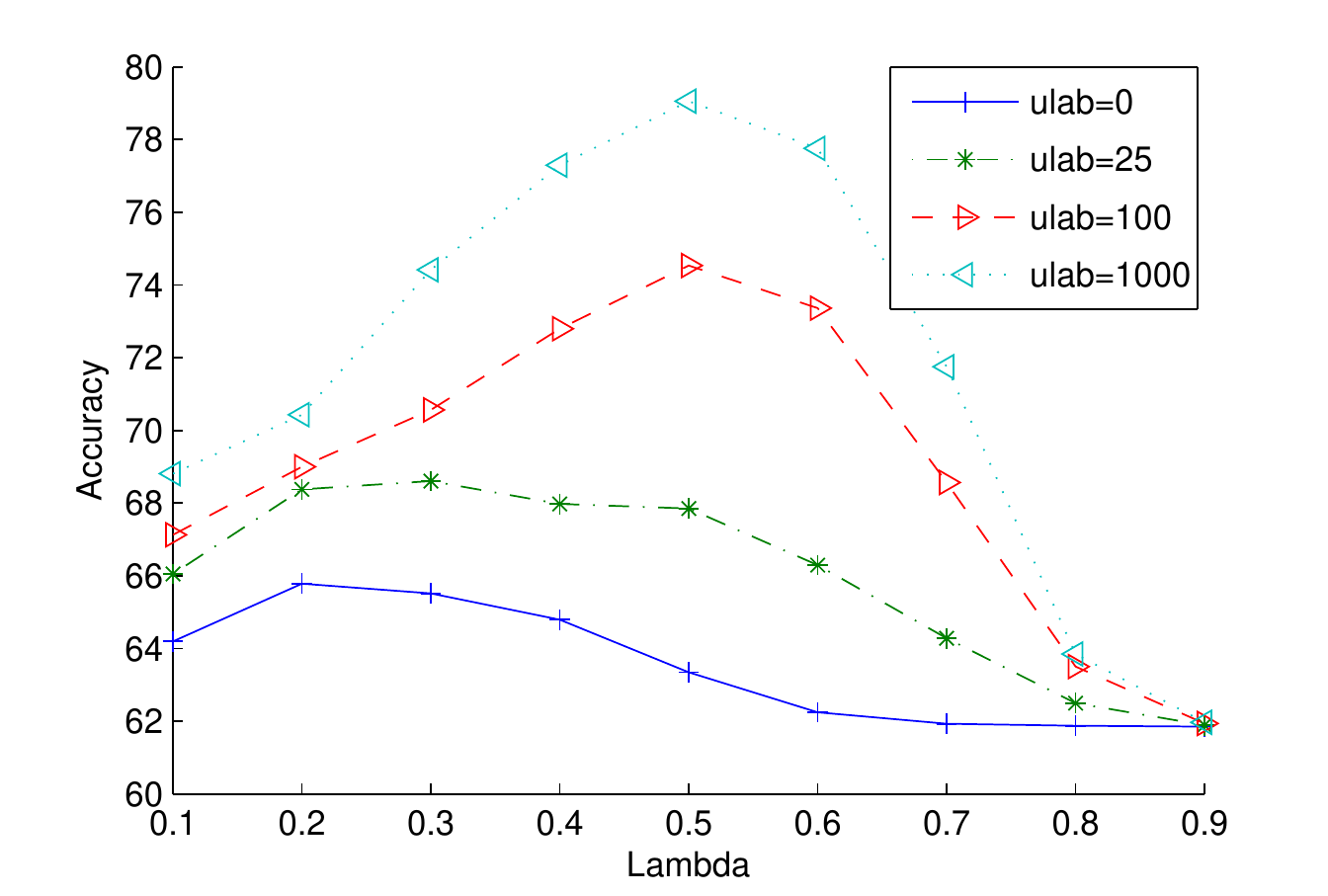}
\caption{Results for sraa dataset for different number of unlabeled examples. Number of labeled examples=10.}
\label{fig:sraa}
\end{center}
\end{figure}

\begin{figure}[tc]
\begin{center}
\includegraphics[width=0.5\textwidth]{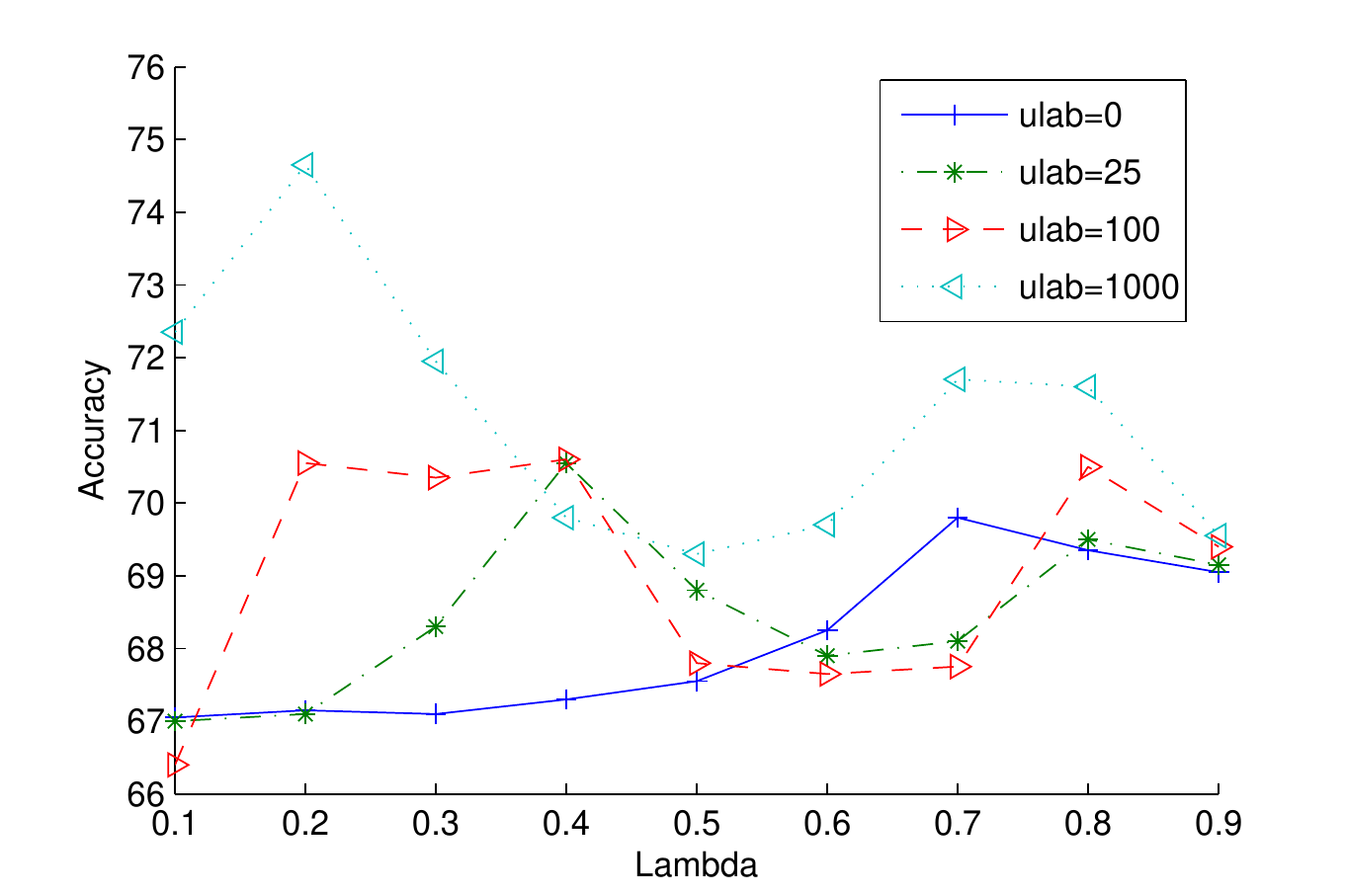}
\caption{Results for movie dataset for different number of unlabeled examples, Number of labeled examples=25.}
\label{fig:movie}
\end{center}
\end{figure}

\figref{sraa} and \figref{movie} show the results for accuracy vs. $\lambda$ for different number of unlabeled examples for {\it sraa} and {\it movie} datasets respectively.  Remember that $\lambda = 0$ is the purely generative model and $\lambda = 1$ is the purely discriminative model.  In both of these figures, we see that as we increase the number of unlabeled examples, performance improves. In {\it sraa}, we observe that increasing the number of unlabeled examples results in the shifting of optimal $\lambda$ ($\lambda^*$) towards rights.  We get an optimal $\lambda^*=0.2$ for a fully supervised model while for $1000$ unlabeled examples, we get $\lambda^*=0.5$.  All the curves in this experiment are uni-modal which means that there is a unique value of $\lambda$ where hybrid model performs best.

Unfortunately, these nearly-perfectly shaped curves are not common to all settings.  We do not observe it in the other dataset (\figref{movie}).  There are values of $\lambda$ where a fully supervised model performs better than the best semi-supervised model.  This experiment emphasizes the need for choosing the right value of $\lambda$ and also shows the importance of the hybrid model.  If we do not choose the right value of $\lambda$, we might end up hurting the model by using the unlabeled data.  We also observe that {\it movie} dataset gives us a bi-modal curve in contrast to the uni-modal curve obtained in the {\it sraa}.  We see that curve is a uni-modal in the supervised setting but as we introduce unlabeled examples, the curves not only become bi-modal but also shift towards the left-hand side (best accuracy is achieved close to the generative end).  This naturally suggests that generative model is actually affecting the hybrid model in a positive manner and exploiting the strength of the unlabeled examples.

\section{Conclusion and Future Work}
We have presented a generalized ``PCP'' hybrid model for the exponential family distributions and have experimentally shown that the prior conjugate to the generative model is more appropriate than the Gaussian prior.  In addition to the performance advantage, the conjugate prior also gives us a closed form solution for the generative parameters.  In the future, we aim at interpreting these results in a theoretical way and answer questions like: (1) Under what conditions will the hybrid model perform better than both the generative and discriminative models?  (2) What is the optimal value of $\gamma$?  (3) Is a PAC-style analysis of the hybrid model possible for the finite sample case as opposed the asymptotic analysis mostly found in the literature?

{\footnotesize
\bibliography{ref}

\begin{thebibliography}{}

\bibitem[\protect\citeauthoryear{Bouchard and Triggs}{2004}]{Triggs2004}
G.~Bouchard and Bill Triggs.
\newblock The tradeoff between generative and discriminative classifiers.
\newblock In {\em IASC International Symposium on Computational Statistics},
  pages 721--728, Prague, August 2004.

\bibitem[\protect\citeauthoryear{Bouchard}{2007}]{Bouchard2007}
Guillaume Bouchard.
\newblock Bias-variance tradeoff in hybrid generative-discriminative models.
\newblock In {\em Proceedings of the Sixth International Conference on Machine
  Learning and Applications}, pages 124--129, Washington, DC, USA, 2007. IEEE
  Computer Society.

\bibitem[\protect\citeauthoryear{Cozman \bgroup \em et al.\egroup
  }{2003}]{Cozman03}
Fabio~Gagliardi Cozman, Ira Cohen, Marcelo~Cesar Cirelo, and Escola
  Politécnica.
\newblock Semi-supervised learning of mixture models.
\newblock In {\em 20th International Conference on Machine Learning}, pages
  99--106, 2003.

\bibitem[\protect\citeauthoryear{Druck \bgroup \em et al.\egroup
  }{2007}]{Druck2007}
Gregory Druck, Chris Pal, Andrew {McCallum}, and Xiaojin Zhu.
\newblock Semi-supervised classification with hybrid generative/discriminative
  methods.
\newblock In {\em Proceedings of the 13th ACM SIGKDD international conference
  on Knowledge discovery and data mining}, pages 280--289, New York, NY, USA,
  2007. ACM.

\bibitem[\protect\citeauthoryear{Fujino \bgroup \em et al.\egroup
  }{2007}]{Fujino2008}
Akinori Fujino, Naonori Ueda, and Kazumi Saito.
\newblock A hybrid generative/discriminative approach to text classification
  with additional information.
\newblock {\em Inf. Process. Manage.}, 43(2):379--392, 2007.

\bibitem[\protect\citeauthoryear{Jaakkola and Haussler}{1999}]{Jaakkola99}
Tommi~S. Jaakkola and David Haussler.
\newblock Exploiting generative models in discriminative classifiers.
\newblock In {\em Advances in Neural Information Processing Systems 11}, pages
  487--493. MIT Press, 1999.

\bibitem[\protect\citeauthoryear{Lasserre \bgroup \em et al.\egroup
  }{2006}]{Minka2006}
Julia~A. Lasserre, Christopher~M. Bishop, and Thomas~P. Minka.
\newblock Principled hybrids of generative and discriminative models.
\newblock In {\em Proceedings of the 2006 IEEE Computer Society Conference on
  Computer Vision and Pattern Recognition}, pages 87--94, Washington, DC, USA,
  2006. IEEE Computer Society.

\bibitem[\protect\citeauthoryear{Mccallum and Nigam}{1998}]{Mccallum1998}
A.~Mccallum and K.~Nigam.
\newblock A comparison of event models for naive bayes text classification.
\newblock In {\em AAAI Workshop on "Learning for Text Categorization"}, 1998.

\bibitem[\protect\citeauthoryear{{McCallum} \bgroup \em et al.\egroup
  }{2006}]{Mccallum06}
Andrew {McCallum}, Chris Pal, Greg Druck, and Xuerui Wang.
\newblock Multi-conditional learning: Generative/discriminative training for
  clustering and classification.
\newblock In {\em Proceedings of the 21st National Conference on Artificial
  Intelligence}, pages 433--439, 2006.

\bibitem[\protect\citeauthoryear{Ng and Jordan}{2002}]{Ng2002}
Andrew~Y. Ng and Michael~I. Jordan.
\newblock On discriminative vs. generative classifiers: A comparison of
  logistic regression and naive bayes.
\newblock In {\em Advances in Neural Information Processing Systems 14},
  Cambridge, MA, 2002. MIT Press.

\bibitem[\protect\citeauthoryear{Nigam \bgroup \em et al.\egroup
  }{2000}]{Nigam2000}
Kamal Nigam, Andrew~K. {McCallum}, Sebastian Thrun, and Tom Mitchell.
\newblock Text classification from labeled and unlabeled documents using em.
\newblock {\em Machine Learning}, V39(2):103--134, May 2000.

\bibitem[\protect\citeauthoryear{Raina \bgroup \em et al.\egroup
  }{2003}]{Raina2003}
Rajat Raina, Yirong Shen, Andrew~Y. Ng, and Andrew {McCallum}.
\newblock Classification with hybrid generative/discriminative models.
\newblock In {\em Advances in Neural Information Processing Systems 16}. MIT
  Press, 2003.

\bibitem[\protect\citeauthoryear{Zhu}{2005}]{zhu05survey}
Xiaojin Zhu.
\newblock Semi-supervised learning literature survey.
\newblock Technical Report 1530, Computer Sciences, University of
  Wisconsin-Madison, 2005.

\end{thebibliography}
\bibliographystyle{named}
}

\end{document}